\documentclass[11pt,a4paper]{article}
\usepackage[hyperref]{emnlp-ijcnlp-2019}
\usepackage{times}
\usepackage{soul}
\usepackage{url}
\usepackage[utf8]{inputenc}
\usepackage{latexsym}
\usepackage{graphicx}
\usepackage{amsmath}
\usepackage{booktabs}
\usepackage{algorithm}
\usepackage{algorithmic}
\usepackage{makecell}
\usepackage{multirow}
\usepackage{rotating}
\usepackage{subcaption}
\newcommand{\hlc}[2][yellow]{{%
    \colorlet{foo}{#1}%
    \sethlcolor{foo}\hl{#2}}%
}

\usepackage{url}

\aclfinalcopy % Uncomment this line for the final submission

\title{Uncover Sexual Harassment Patterns from Personal Stories by Joint Key Element Extraction and Categorization}

\author{Yingchi Liu$^1$, Quanzhi Li$^1$, Marika Cifor$^2$, Xiaozhong Liu$^3$, Qiong Zhang$^1$ and Luo Si$^1$ \\
  $^1$Alibaba Group, US\\
  $^2$University of Washington, USA\\
  $^3$Indiana University Bloomington, USA\\
  {\tt \{yingchi.liu, quanzhi.li, qz.zhang,luo.si\}@alibaba-inc.com} \\
    {\tt mcifor@uw.edu}, {\tt liu237@indiana.edu} \\}

\date{}

\begin{document}
\maketitle
\begin{abstract}
The number of personal stories about sexual harassment shared online has increased exponentially in recent years. This is in part inspired by the \#MeToo and \#TimesUp movements. Safecity is an online forum for people who experienced or witnessed sexual harassment to share their personal experiences. It has collected \textgreater 10,000 stories so far. Sexual harassment occurred in a variety of situations, and categorization of the stories and extraction of their key elements will provide great help for the related parties to understand and address sexual harassment. In this study, we manually annotated those stories with labels in the dimensions of location, time, and harassers' characteristics, and marked the key elements related to these dimensions. Furthermore, we applied natural language processing technologies with joint learning schemes to automatically categorize these stories in those dimensions and extract key elements at the same time. We also uncovered significant patterns from the categorized sexual harassment stories. We believe our annotated data set, proposed algorithms, and analysis will help people who have been harassed, authorities, researchers and other related parties in various ways, such as automatically filling reports, enlightening the public in order to prevent future harassment, and enabling more effective, faster action to be taken.
\end{abstract}

\section{Introduction}
Sexual violence, including harassment, is a pervasive, worldwide problem with a long history. This global problem has finally become a mainstream issue thanks to the efforts of survivors and advocates. Statistics show that girls and women are put at high risk of experiencing harassment. Women have about a 3 in 5 chance of experiencing sexual harassment, whereas men have slightly less than 1 in 5 chance \cite{web1,web2,web7}. While women in developing countries are facing distinct challenges with sexual violence \cite{Lea2017}, however sexual violence is ubiquitous. In the United States, for example, there are on average \textgreater 300,000 people who are sexually assaulted every year  \cite{DOJ2017}. Additionally, these numbers could be underestimated, due to reasons like guilt, blame, doubt and fear, which stopped many survivors from reporting \cite{web3}. Social media can be a more open and accessible channel for those who have experienced harassment to be empowered to freely share their traumatic experiences and to raise awareness of the vast scale of sexual harassment, which then allows us to understand and actively address abusive behavior as part of larger efforts to prevent future sexual harassment. The deadly gang rape of a medical student on a Delhi bus in 2012 was a catalyst for protest and action, including the development of Safecity, which uses online and mobile technology to work towards ending sexual harassment and assault. More recently, the \#MeToo and \#TimesUp movements, further demonstrate how reporting personal stories on social media can raise awareness and empower women. Millions of people around the world have come forward and shared their stories. Instead of being bystanders, more and more people become up-standers, who take action to protest against sexual harassment online. The stories of people who experienced harassment can be studied to identify different patterns of sexual harassment, which can enable solutions to be developed to make streets safer and to keep women and girls more secure when navigating city spaces \cite{karlekar2018safecity}. In this paper, we demonstrated the application of natural language processing (NLP) technologies to uncover harassment patterns from social media data.  We made three key contributions: 

1. Safecity\footnote{\url{https://safecity.in}} is the largest publicly-available online forum for reporting sexual harassment \cite{karlekar2018safecity}. We annotated about 10,000 personal stories from Safecity with the key elements, including information of harasser (i.e. the words describing the harasser), time, location and the trigger words (i.e. the phrases indicate the harassment that occurred). The key elements are important for studying the patterns of harassment and victimology \cite{web3,Ceccato2017}. Furthermore, we also associated each story with five labels that characterize the story in multiple dimensions (i.e. age of harasser, single/multiple harasser(s), type of harasser, type of location and time of day). The annotation data are available online.\footnote{\url{https://github.com/alievent/harassment-analysis} Please follow Safecity guidelines for usage.} 

2. We proposed joint learning NLP models that use convolutional neural network (CNN) \cite{LeCun:1998} and bi-directional long short-term memory (BiLSTM)  \cite{Schuster97bidirectionalrecurrent,doi:10.1162/neco.1997.9.8.1735} as basic units. Our models can automatically extract the key elements from the sexual harassment stories and at the same time categorize the stories in different dimensions. The proposed models outperformed the single task models, and achieved higher than previously reported accuracy in classifications of harassment forms \cite{karlekar2018safecity}.

3. We uncovered significant patterns from the categorized sexual harassment stories.

\section{Related Work}
Conventional surveys and reports are often used to study sexual harassment, but harassment on these is usually under-reported \cite{web7,web3}. The high volume of social media data available online can provide us a much larger collection of firsthand stories of sexual harassment. Social media data has already been used to analyze and predict distinct societal and health issues, in order to improve the understanding of wide-reaching societal concerns, including mental health, detecting domestic abuse, and cyberbullying \cite{Balani:2015,D15-1309,Ziegele2018,DBLP:journals/corr/abs-1801-06482}. 

There are a very limited number of studies on sexual harassment stories shared online. Karlekar and Bansal \shortcite{karlekar2018safecity} were the first group to our knowledge that applied NLP to analyze large amount ( $\sim$10,000) of sexual harassment stories. Although their CNN-RNN classification models demonstrated high performance on classifying the forms of harassment, only the top 3 majority forms were studied. In order to study the details of the sexual harassment, the trigger words are crucial. Additionally, research indicated that both situational factors and person (or individual difference) factors contribute to sexual harassment \cite{Hitlan2009}. Therefore, the information about perpetrators needs to be extracted as well as the location and time of events. Karlekar and Bansal \shortcite{karlekar2018safecity} applied several visualization techniques in order to capture such information, but it was not obtained explicitly. Our preliminary research demonstrated automatic extraction of key element and story classification in separate steps \cite{Liu2019}. In this paper, we proposed joint learning NLP models to directly extract the information of the harasser, time, location and trigger word as key elements and categorize the harassment stories in five dimensions as well. Our approach can provide an avenue to automatically uncover nuanced circumstances informing sexual harassment from online stories. 
\section{Data Collection and Annotation}
 \begin{figure}[t]
% \vspace{-5pt}
    \includegraphics[width=7.7 cm, height=0.72cm]{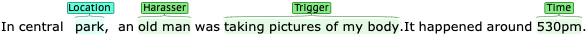}
    \caption{An example of the annotation.}
  \label{fig:ann}
  \vspace{-10pt}
\end{figure}

We obtained 9,892 stories of sexual harassment incidents that was reported on Safecity. Those stories include a text description, along with tags of the forms of harassment, e.g. commenting, ogling and groping. A dataset of these stories was published by Karlekar and Bansal \shortcite{karlekar2018safecity}. In addition to the forms of harassment, we manually annotated each story with the key elements (i.e. ``harasser", ``time", ``location", ``trigger"), because they are essential to uncover the harassment patterns. An example is shown in Figure \ref{fig:ann}. Furthermore, we also assigned each story classification labels in five dimensions (Table \ref{tab:classDef}). The detailed definitions of classifications in all dimensions are explained below.
  \begin{table*}[t]
%   \vspace{-5pt}
  \centering
 \small
  \begin{tabular}{l|l}
    \toprule
    Dimensions & Classes/Groups\\
    \midrule
    Age of Harasser& 0 - unspecified,  1 - teenagers or young people, 2 - adults \\
        \midrule
    Single/Multiple Harasser(s)&0 - unspecified, 1 - single harasser, 2 - multiple harassers\\
        \midrule
    Type of Harasser&0 - unspecified, 1 - relative, 2 - teacher, 3 - classmate, 4 - friend, 5 - neighbor,\\
    & 6 - conductor/driver, 7 - work-related, 8 - police/guard, 9 - other\\
        \midrule
    Type of Location & 0 - unspecified, 1 - street, 2 - transportation, 3 - station/stop, 4 - private places,\\
    & 5 - shopping places, 6 - neighborhood, 7 - park, 8 - hotel, 9 - bush/woods, \\
    &10 - parking lot, 11 - in/near school, 12 - restaurant, 13 - other\\
        \midrule
    Time of Day& 0 - unspecified, 1 - day (5am to 6pm), 2 - evening or night\\
  \bottomrule
\end{tabular}
  \caption{Definition of classes in different dimensions about sexual harassment.}
 \vspace{-10pt}
\label{tab:classDef}
\end{table*}

{\bf Age of Harasser}: Individual difference such as age can affect harassment behaviors. Therefore, we studied the harassers in two age groups, young and adult. Young people in this paper refer to people in the early 20s or younger.

{\bf Single/Multiple Harasser(s)}: Harassers may behave differently in groups than they do alone. 

{\bf Type of Harasser}: Person factors in harassment include the common relationships or titles of the harassers. Additionally, the reactions of people who experience harassment may vary with the harassers' relations to themselves \cite{web3}. We defined 10 groups with respects to the harassers' relationships or titles. We put conductors and drivers in one group, as they both work on the public transportation. Police and guards are put in the same category, because they are employed to provide security. Manager, supervisors, and colleagues are in the work-related group. The others are described by their names.

{\bf Type of Location}: It will be helpful to reveal the places where harassment most frequently occurs \cite{Ceccato2017,karlekar2018safecity}. We defined 14 types of locations. ``Station/stop'' refers to places where people wait for public transportation or buy tickets. Private places include survivors' or harassers' home, places of parties and etc. The others are described by their names.

{\bf Time of Day}: The time of an incident may be reported as \textit{``in evening”} or at a specific time, e.g. \textit{``10 pm''}. We considered that 5 am to 6 pm as day time, and the rest of the day as the night.

Because many of the stories collected are short, many do not contain all of the key elements. For example, \textit{``A man came near to her tried to be physical with her .''}. The time and location are unknown from the story. In addition, the harassers were strangers to those they harassed in many cases. For instance, \textit{``My friend was standing in the queue to pay bill and was ogled by a group of boys.''}, we can only learn that there were multiple young harassers, but the type of harasser is unclear. The missing information is hence marked as ``unspecified''. It is different from the label ``other", which means the information is provided but the number of them is too small to be represented by a group, for example, a ``trader''. 

All the data were labeled by two annotators with training. Inter-rater agreement was measured by Cohen's kappa coefficient, ranging from 0.71 to 0.91 for classifications in different dimensions and 0.75 for key element extraction (details can refer to Table 1 in supplementary file). The disagreements were reviewed by a third annotator and a final decision was made. 

\section{Proposed Models}
The key elements can be very informative when categorizing the incidents. For instance, in Figure 1, with identified key elements, one can easily categorize the incident in dimensions of ``age of harasser'' (adult), ``single/multiple harasser(s)'' (single), ``type of harasser'' (unspecified), ``type of location'' (park) , ``time of day'' (day time). Therefore, we proposed two joint learning schemes to extract the key elements and categorize the incidents together. In the models' names, ``J'', ``A'', ``SA'' stand for joint learning, attention, and supervised attention, respectively.

\subsection{CNN Based Joint Learning Models}
\begin{figure*}[t]
\centering
  \includegraphics[width=15cm,height=6.8cm]{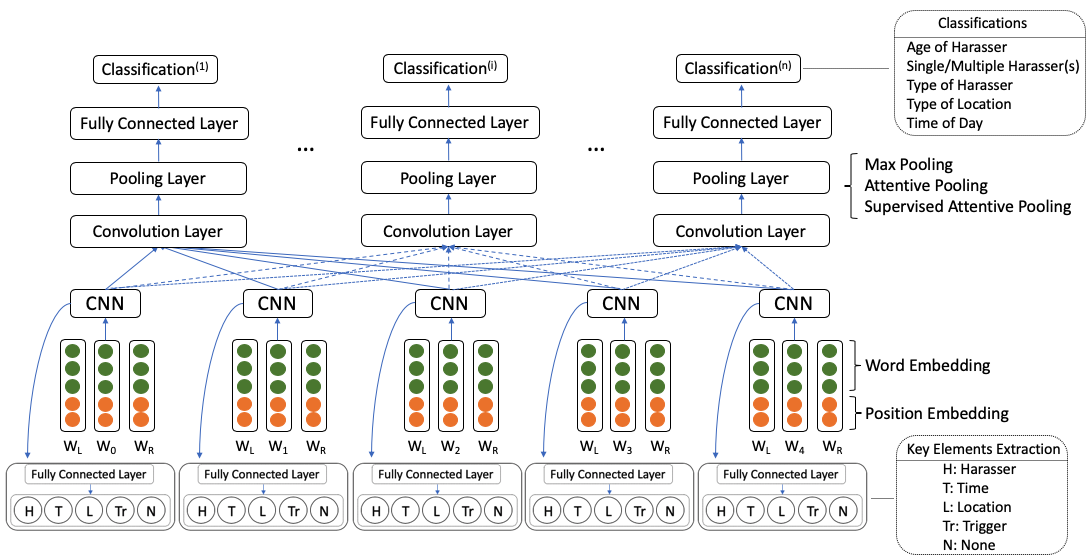}\hfill
  \caption{CNN based Joint learning Model. $W_{L}$ and $W_{R}$ are the left and right context around each word. }
  \label{fig:cnn}
  %\vspace{-10pt}
  \end{figure*}
\begin{figure*}[t]
\centering
  \includegraphics[width=14.8cm,height=6.5cm]{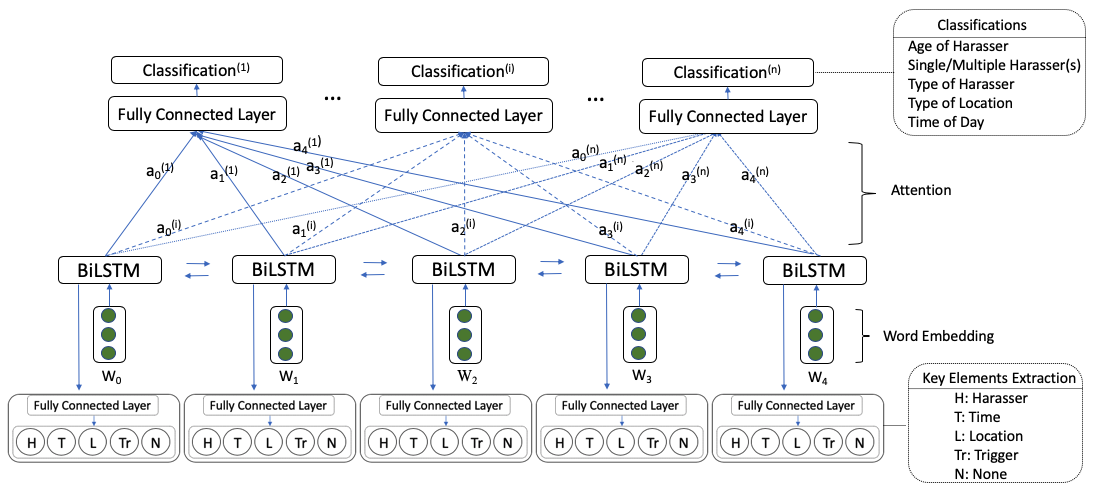}\hfill
  \caption{BiLSM based Joint Learning Model. Here we use an input of five words as an example.}
  \label{fig:bilstm}
   %\vspac e{-10pt}
 \end{figure*}
In Figure \ref{fig:cnn}, the first proposed structure consists of two layers of CNN modules. 

{\bf J-CNN:} To predict the type of key element, it is essential for the CNN model to capture the context information around each word. Therefore, the word along with its surrounding context of a fixed window size was converted into a context sequence. Assuming a window size of $2l + 1$ around the target word $w_0$, the context sequence is $[(w_{-l}, w_{-l+1},...w_0, ...w_{l-1},w_l)]$, where $w_i (i \in[-l,l])$ stands for the $ith$ word from $w_0$.

Because the context of the two consecutive words in the original text are only off by one position, it will be difficult for the CNN model to detect the difference. Therefore, the position of each word in this context sequence is crucial information for the CNN model to make the correct predictions \cite{Nguyen2015EventDA}. That position was embedded as a $p$ dimensional vector, where $p$ is a hyperparameter. The position embeddings were learned at the training stage. Each word in the original text was then converted into a sequence of the concatenation of word and position embeddings. Such sequence was fed into the CNN modules in the first layer of the model, which output the high level word representation ($h_i, i\in[0,n-1]$, where n is the number of input words). The high level word representation was then passed into a fully connected layer, to predict the key element type for the word. The CNN modules in this layer share the same parameters.

We input the sequence of high level word representations ($h_i$) from the first layer into another layer of multiple CNN modules to categorize the harassment incident in each dimension (Figure \ref{fig:cnn}). Inside each CNN module, the sequence of word representations were first passed through a convolution layer to generate a sequence of new feature vectors ($C =[c_0,c_1,...c_q]$). This vector sequence ($C$) was then fed into a max pooling layer. This is followed by a fully connected layer. Modules in this layer do not share parameters across classification tasks.

{\bf J-ACNN:} We also experimented with attentive pooling, by replacing the max pooling layer. The attention layer aggregates the sequence of feature vectors ($C$) by measuring the contribution of each vector to form the high level representation of the harassment story. Specifically,
\begin{align}
u_{i} = tanh(W_{\omega}c_{i} + b_{\omega})
\end{align}
\begin{align}
\alpha_{i} = \frac{exp(u^{T}_{i}u_{\omega})}{\sum_{i}exp(u^{T}_{i}u_{w})}
\end{align}
\begin{align}
v=\sum_{i}\alpha_i c_{i}
\end{align}

That is, a fully connected layer with non-linear activation was applied to each vector $c_{i}$ to get its hidden representation $u_{i}$. The similarity of $u_{i}$ with a context vector $u_{w}$ was measured and get normalized through a softmax function, as the importance weight $\alpha_{i}$. The final representation of the incident story $v$ was an aggregation of all the feature vectors weighted by $\alpha_{i}$. $W_{\omega}$, $b_{\omega}$ and $u_{w}$ were learned during training.

The final representation ($v$) was passed into one fully connected layer for each classification task. We also applied different attention layers for different classifications, because the classification modules categorize the incident in different dimensions, their focuses vary. For example, to classify ``time of day'', one needs to focus on the time phrases, but pays more attention to harassers when classifying ``age of harasser''.

{\bf J-SACNN:} To further exploit the information of the key elements, we applied supervision \cite{P18-2066} to the attentive pooling layer, with the annotated key element types of the words as ground truth. For instance, in classification of ``age of harasser'', the ground truth attention labels for words with key element types of ``harasser'' are 1 and others are 0. To conform to the CNN structure, we applied convolution to the sequence of ground truth attention labels, with the same window size ($w$) that was applied to the word sequence (Eq. \ref{eq:attention convolution}).
\begin{align}
    \alpha^{*}(t) = W \circ [e_t : e_{t + w -1}] 
    \label{eq:attention convolution}
\end{align}
where $\circ$ is element-wise multiplication,  $e_t$ is the ground truth attention label, and the $W \in R^{w\times1}$ is a constant matrix with all elements equal to 1. $\alpha^{*}$ was normalized through a softmax function and used as ground truth weight values of the vector sequence ($C$) output from the convolution layer. The loss was calculated between learned attention $\alpha$ and $\alpha^{*}$ (Eq. \ref{eq:attention loss}), and added to the total loss.
\begin{align}
E(\alpha,\alpha^{*}) = \sum^{q}_{i=0}(\alpha_{i}-\alpha^{*}_{i})^{2}
\label{eq:attention loss}
\end{align}

\subsection{BiLSTM Based Joint Learning Models}
{\bf J-BiLSTM:} The model input the sequence of word embeddings to the BiLSTM layer. To extract key elements, the hidden states from the forward and backward LSTM cells were concatenated and used as word representations to predict the key element types. 

To classify the harassment story in different dimensions, concatenation of the forward and backward final states of BiLSTM layer was used as document level representation of the story.

{\bf J-ABiLSTM:} We also experimented on BiLSTM model with the attention layer to aggregate the outputs from BiLSTM layer (Figure \ref{fig:bilstm}). The aggregation of the outputs was used as document level representation.

{\bf J-SABiLSTM:} Similarly, we experimented with the supervised attention. 

In all the models, softmax function was used to calculate the probabilities at the prediction step, and the cross entropy losses from extraction and classification tasks were added together. In case of supervised attention, the loss defined in Eq. \ref{eq:attention loss} was added to the total loss as well. We applied the stochastic gradient descent algorithm with mini-batches and the AdaDelta update
Rule (rho=0.95 and epsilon=1e-6) \cite{Zeiler2012,Feng2016}. The gradients were computed using back-propagation. During training, we also optimized the word and position embeddings. 

\section{Experiments and Results}
\subsection{Experimental Settings}
{\bf Data Splits}: We used the same splits of train, develop, and test sets used by Karlekar and Bansal \cite{karlekar2018safecity}, with 7201, 990 and 1701 stories, respectively. In  this  study,  we  only  considered single label classifications.

{\bf Baseline Models}: CNN and BiLSTM models that perform classification and extraction separately were used as baseline models. In classification, we also experimented with BiLSTM with the attention layer. To demonstrate that the improvement came from joint learning structure rather the two layer structure in J-CNN, we investigated the same model structure without training on key element extraction. We use J-CNN* to denote it. 

{\bf Preprocess}: All the texts were converted to lowercase and preprocessed by removing non-alphanumeric characters, excluding ``. ! ? '' . The word embeddings were pre-trained using fastText \cite{bojanowski2016enriching} with dimension equaling 100. 

{\bf Hyperparameters}: For the CNN model, the filter size was chosen to be (1,2,3,4), with 50 filters per filter size. Batch size was set to 50 and the dropout rate was 0.5. The BiLSTM model comprises two layers of one directional LSTM. Every LSTM cell has 50 hidden units. The dropout rate was 0.25. Attention size was 50.

\subsection{Results and Discussions}
We compared joint learning models with the single task models. Results are averages from five experiments. Although not much improvement was achieved in key element extraction (Figure \ref{tab:extractionRes}), classification performance improved significantly with joint learning schemes (Table \ref{tab:cls}). Significance t-test results are shown in Table 2 in the supplementary file.

{\bf BiLSTM Based Models:} Joint learning BiLSTM with attention outperformed single task BiLSTM models. One reason is that it directed the attention of the model to the correct part of the text. For example,

\textbf{S1:} \textit{``\hlc[green!1.7003483371809125]{when} \hlc[green!3.4324652515351772]{i} \hlc[green!10.76661329716444]{was} \hlc[green!20.388443022966385]{returning} \hlc[green!9.704475291073322]{my} \hlc[green!6.052316632121801]{home} \hlc[green!2.477810252457857]{after} \hlc[green!3.5612427163869143]{finishing} \hlc[green!4.7736018896102905]{my} \hlc[green!4.634172189980745]{class} \hlc[green!0.6899426807649434]{.} \hlc[green!0.35572052001953125]{i} \hlc[green!0.3427551419008523]{was} \hlc[green!0.293194578262046]{in} \hlc[green!0.2028885210165754]{queue} \hlc[green!0.10553237370913848]{to} \hlc[green!0.19472737039905041]{get} \hlc[green!0.44946340494789183]{on} \hlc[green!0.5511227645911276]{the} \hlc[green!2.056689700111747]{micro} \hlc[green!2.597035141661763]{bus} \hlc[green!2.5683704297989607]{and} \hlc[green!4.6382867731153965]{there} \hlc[green!9.827975183725357]{was} \hlc[green!21.346069872379303]{a} \hlc[green!22.295180708169937]{girl} \hlc[green!11.672522872686386]{opposite} \hlc[green!8.892465382814407]{to} \hlc[green!18.20233091711998]{me} \hlc[green!13.192926533520222]{just} \hlc[green!26.24184638261795]{then} \hlc[green!40.2555949985981]{a} \hlc[green!30.108729377388954]{young} \hlc[green!115.02625793218613]{man} \hlc[green!93.40204298496246]{tried} \hlc[green!58.68498980998993]{to} \hlc[green!144.01434361934662]{touch} \hlc[green!108.82275551557541]{her} \hlc[green!80.9452086687088]{on} \hlc[green!47.26015031337738]{the} \hlc[green!47.71501570940018]{breast} \hlc[green!19.392695277929306]{.}''}

\textbf{S2:} \textit{``\hlc[green!0.2212507533840835]{when} \hlc[green!0.26129744946956635]{i} \hlc[green!0.3014186804648489]{was} \hlc[green!0.314583390718326]{returning} \hlc[green!0.23829322890378535]{my} \hlc[green!0.018542312318459153]{home} \hlc[green!0.06052045864635147]{after} \hlc[green!0.3865368489641696]{finishing} \hlc[green!0.5127551266923547]{my} \hlc[green!0.569560332223773]{class} \hlc[green!0.037081812479300424]{.} \hlc[green!0.061129467212595046]{i} \hlc[green!0.12043083552271128]{was} \hlc[green!0.2053432835964486]{in} \hlc[green!0.038308095099637285]{queue} \hlc[green!0.05270353358355351]{to} \hlc[green!0.07939991337480024]{get} \hlc[green!0.14962266141083091]{on} \hlc[green!0.11444976553320885]{the} \hlc[green!0.013002995729038958]{micro} \hlc[green!0.016201976904994808]{bus} \hlc[green!0.14046543219592422]{and} \hlc[green!0.12413455988280475]{there} \hlc[green!0.18423641449771821]{was} \hlc[green!0.3394613158889115]{a} \hlc[green!1.0372470133006573]{girl} \hlc[green!0.20553644571918994]{opposite} \hlc[green!0.2821453963406384]{to} \hlc[green!0.5574009846895933]{me} \hlc[green!0.2709480468183756]{just} \hlc[green!0.2582515007816255]{then} \hlc[green!0.9223996312357485]{a} \hlc[green!788.9420390129089]{young} \hlc[green!199.1765946149826]{man} \hlc[green!0.39259070763364434]{tried} \hlc[green!0.27069455245509744]{to} \hlc[green!0.5092779756523669]{touch} \hlc[green!0.7033208385109901]{her} \hlc[green!0.6793316570110619]{on} \hlc[green!0.5892394692637026]{the} \hlc[green!0.4084075626451522]{breast} \hlc[green!0.14951340563129634]{.}''}

\textbf{S3:} \textit{``\hlc[green!0.23944019631017]{when} \hlc[green!0.16698541003279388]{i} \hlc[green!0.3381385176908225]{was} \hlc[green!0.21315943740773946]{returning} \hlc[green!0.3222442464902997]{my} \hlc[green!0.8483575657010078]{home} \hlc[green!0.10339960863348097]{after} \hlc[green!0.2440519310766831]{finishing} \hlc[green!0.39699181797914207]{my} \hlc[green!1.2218113988637924]{class} \hlc[green!0.1232976937899366]{.} \hlc[green!0.10928708070423454]{i} \hlc[green!0.2562549489084631]{was} \hlc[green!0.8099888218566775]{in} \hlc[green!2.9650430660694838]{queue} \hlc[green!0.507337914314121]{to} \hlc[green!0.727736041881144]{get} \hlc[green!0.7367140497080982]{on} \hlc[green!0.711284636054188]{the} \hlc[green!194.2763775587082]{micro} \hlc[green!786.8869304656982]{bus} \hlc[green!0.4422159108798951]{and} \hlc[green!0.43104542419314384]{there} \hlc[green!0.4694198723882437]{was} \hlc[green!0.5085613229312003]{a} \hlc[green!0.4430979897733778]{girl} \hlc[green!0.36199347232468426]{opposite} \hlc[green!0.31067250529304147]{to} \hlc[green!0.2927705936599523]{me} \hlc[green!0.24646619567647576]{just} \hlc[green!0.23911069729365408]{then} \hlc[green!0.11775700113503262]{a} \hlc[green!0.002219072712250636]{young} \hlc[green!0.0019248132048232947]{man} \hlc[green!0.32698659924790263]{tried} \hlc[green!0.3118939639534801]{to} \hlc[green!0.5727249081246555]{touch} \hlc[green!0.5670131067745388]{her} \hlc[green!0.7104063988663256]{on} \hlc[green!0.6698771030642092]{the} \hlc[green!0.4756081907544285]{breast} \hlc[green!0.26600153069011867]{.}''}

In S1, the regular BiLSTM with attention model for classification on ``age of harasser'' put some attention on phrases other than the harasser, and hence aggregated noise. This could explain why the regular BiLSTM model got lower performance than the CNN model. However, when training with key element extractions, it put almost all attention on the harasser ``young man'' (S2), which helped the model make correct prediction of ``young harasser''. When predicting the ``type of location'' (S3), the joint learning model directed its attention to ``micro bus''.
\begin{table}[tb]
\centering
\small
\begin{tabular}{l|r|r}
\toprule
&Accuracy&Macro F1\\
\midrule
BiLSTM&92.1&79.9\\
CNN&\textbf{92.6}&79.5\\
\midrule
J-BiLSTM&92.1&79.2\\
J-ABiLSTM&92.0&79.7\\
J-SABiLSTM&92.0&79.1\\
J-CNN&92.5&80.1\\
J-ACNN&92.4&\textbf{80.4}\\
J-SACNN&92.4&80.1\\
\bottomrule
\end{tabular}
\caption{Key element extraction results.}
\label{tab:extractionRes}
\vspace{-10pt}
\end{table}
\begin{table}[tb]
\centering
\small
  \begin{tabular}{ll@{\hspace{0.2ex}}|r@{\hspace{0.8ex}}r@{\hspace{0.8ex}}r@{\hspace{0.9ex}}r@{\hspace{0.6ex}}r}
  \toprule
    &&  \makecell{Age}   &\makecell{Single/\\Multiple}& \makecell{Type of\\Harasser} & \makecell{Loc} & \makecell{Time}\\
        \midrule
    \multirow{6}{*}{\begin{sideways}Accuracy\end{sideways}}&\multicolumn{1}{|l|}{BiLSTM}&90.7&91.1&91.0&80.3&97.0\\
    &\multicolumn{1}{|l|}{ABiLSTM}&90.0&91.8&91.4&81.6&97.0\\
    &\multicolumn{1}{|l|}{CNN}&91.6&92.8&93.2&83.6&97.4\\
    &\multicolumn{1}{|l|}{J-CNN*}&91.5&	92.7&	92.3&	82.6&	97.2\\
    \cmidrule{2-7}
    &\multicolumn{1}{|l|}{J-BiLSTM}&90.8&91.8&91.3&78.3&94.8\\
    &\multicolumn{1}{|l|}{J-ABiLSTM}&92.4&\textbf{94.0}&92.3&\textbf{85.1}&97.7\\
    &\multicolumn{1}{|l|}{J-SABiLSTM}&92.4&93.7&92.1&84.8&97.4\\
    &\multicolumn{1}{|l|}{J-CNN}&92.5&93.8&\textbf{93.3}&84.2&98.0\\
     &\multicolumn{1}{|l|}{J-ACNN}&\textbf{92.8}&93.3&93.1&84.2&97.9\\
      &\multicolumn{1}{|l|}{J-SACNN}&92.5&93.8&92.7&83.1&\textbf{98.0}\\
    \midrule
    \multirow{6}{*}{\begin{sideways}Macro F1\end{sideways}}&\multicolumn{1}{|l|}{BiLSTM}&90.4&90.3&38.6&44.3&93.8\\
    &\multicolumn{1}{|l|}{ABiLSTM}&89.7&	91.0&	33.5&	45.5&	93.6\\
    &\multicolumn{1}{|l|}{CNN}&91.5&	92.1&	46.7&	48.1&	94.4\\
    &\multicolumn{1}{|l|}{J-CNN*}&91.4&	92.0&	46.8&	45.1&	94.5\\
    \cmidrule{2-7}
    &\multicolumn{1}{|l|}{J-BiLSTM}&90.4&	91.1&	34.1&	34.1&	87.7\\
    &\multicolumn{1}{|l|}{J-ABiLSTM}&92.4&	\textbf{93.4}&	37.9&	48.5&	95.1\\
    &\multicolumn{1}{|l|}{J-SABiLSTM}&92.4&	93.0&	37.8&	48.5&	94.4\\
    &\multicolumn{1}{|l|}{J-CNN}&92.4&	93.1&	49.5&	48.6&	\textbf{95.6}\\
    &\multicolumn{1}{|l|}{J-ACNN}&\textbf{92.8}& 92.8&	47.7&	48.5&	95.3\\
    &\multicolumn{1}{|l|}{J-SACNN}&92.4&	\textbf{93.4}&	\textbf{52.8}&	\textbf{50.5}&	95.2\\
  \bottomrule
\end{tabular}
 \caption{Classification accuracy and macro F1 of the models. The best scores are in bold.}
   \label{tab:cls}
     \vspace{-10pt}
\end{table}
\begin{table}[t]
\centering
\small
  \begin{tabular}{l|@{\hspace{0.8ex}}r@{\hspace{0.8ex}}r@{\hspace{0.8ex}}r}
  \toprule
    Models & \makecell{commenting} &\makecell{ogling}&\makecell{groping}\\
    \midrule
    \makecell[l]{CNN*}&80.9&  82.2&  86.0\\
    \makecell[l]{BiLSTM*}& 81.0& 82.2& 86.2\\
     \makecell[l]{CNN+RNN*}&81.6&84.1&86.5\\
    \midrule
        \makecell[l]{J-BiLSTM}&81.7&83.3&87.1\\
    \makecell[l]{J-ABiLSTM}&82.1&83.2&87.9\\
        \makecell[l]{J-SABiLSTM}&82.4&83.1&87.9\\
         \makecell[l]{J-CNN}&81.9&\textbf{84.4}&87.4\\
     \makecell[l]{J-ACNN}&\textbf{82.6}&83.8&88.1\\
      \makecell[l]{J-SACNN}&82.3&83.6&\textbf{88.7}\\
  \bottomrule
\end{tabular}
\caption{Harassment form classification accuracy of models. * Reported by Karlekar and Bansal \shortcite{karlekar2018safecity}}
\label{tab:formcls}
  \vspace{-10pt}
\end{table}

{\bf CNN Based Models:} Since CNN is efficient for capturing the most useful information \cite{P15-1017}, it is quite suitable for the classification tasks in this study. It achieved better performance than the BiLSTM model. The joint learning method boosted the performance even higher. This is because the classifications are related to the extracted key elements, and the word representation learned by the first layer of CNNs (Figure \ref{fig:cnn}) is more informative than word embedding. By plotting of t-SNEs \cite{Maaten2008} of the two kinds of word vectors, we can see the word representations in the joint learning model made the words more separable (Figure 1 in supplementary file). In addition, no improvement was found with the J-CNN* model, which demonstrated the joint learning with extraction is essential for the improvement.

With supervised attentive pooling, the model can get additional knowledge from key element labels. It helped the model in cases when certain location phrases were mentioned but the incidents did not happen at those locations. For instance, \textit{``I was followed on my way home .''}, max pooling will very likely to predict it as ``private places''. But, it is actually unknown. In other cases, with supervised attentive pooling, the model can distinguish ``metro'' and ``metro station'', which are ``transportation'' and ``stop/station'' respectively. Therefore, the model further improved on classifications on ``type of location'' with supervised attention in terms of macro F1. For some tasks, like ``time of day'', there are fewer cases with such disambiguation and hence max pooling worked well. Supervised attention improved macro F1 in location and harasser classifications, because it made more correct predictions in cases that mentioned location and harasser. But the majority did not mention them. Therefore, the accuracy of J-SACNN did not increase, compared with the other models.

{\bf Classification on Harassment Forms: } In Table \ref{tab:formcls}, we also compared the performance of binary classifications on harassment forms with the results reported by Karlekar and Bansal \shortcite{karlekar2018safecity}. Joint learning models achieved higher accuracy. In some harassment stories, the whole text or a span of the text consists of trigger words of multiple forms, such as \textit{``stare, whistles, start to sing, commenting''}. The supervised attention mechanism will force the model to look at all such words rather than just the one related to the harassment form for classification and hence it can introduce noise. This can explain why J-SACNN got lower accuracy in two of the harassment form classifications, compared to J-ACNN. In addition, J-CNN model did best in ``ogling'' classification.

\section{Patterns of Sexual Harassment}
\begin{figure}[t]
    \includegraphics[width=7.7 cm, height=7.5cm]{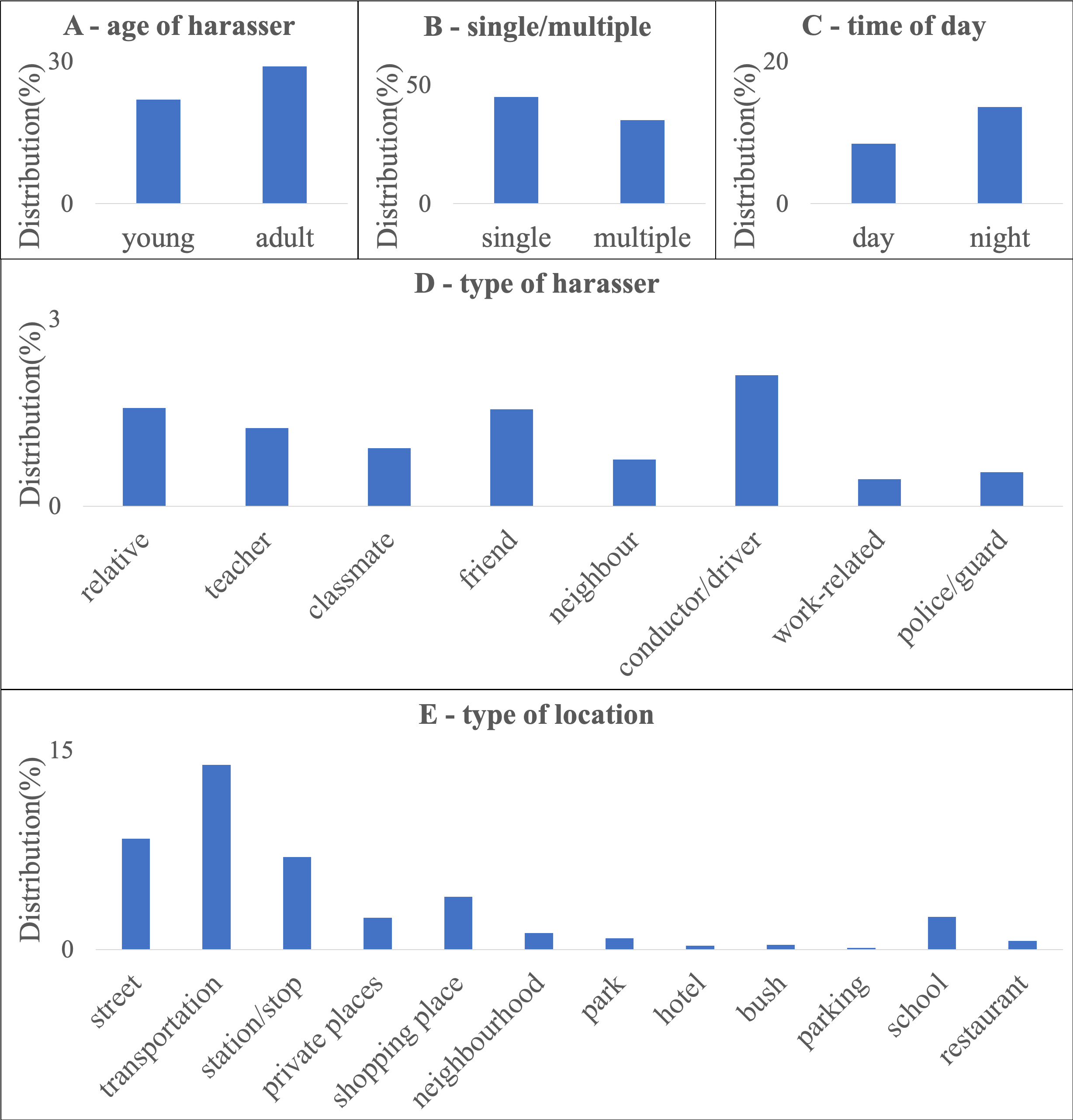}
    \caption{Distributions of incidents.  A) Distributions over age of harasser, B) over single/multiple harasser(s), C) over time of day, D) over  type of harasser. E) over type of location.}
  \label{fig:distr}
    \vspace{-10pt}
\end{figure}

We plotted the distribution of harassment incidents in each categorization dimension (Figure \ref{fig:distr}). It displays statistics that provide important evidence as to the scale of harassment and that can serve as the basis for more effective interventions to be developed by authorities ranging from advocacy organizations to policy makers. It provides evidence to support some commonly assumed factors about harassment: First, we demonstrate that harassment occurred more frequently during the night time than the day time. Second, it shows that besides unspecified strangers (not shown in the figure), conductors and drivers are top the list of identified types of harassers, followed by friends and relatives.
 \begin{figure}[tb]
    \includegraphics[width=7.7 cm, height=6.8cm]{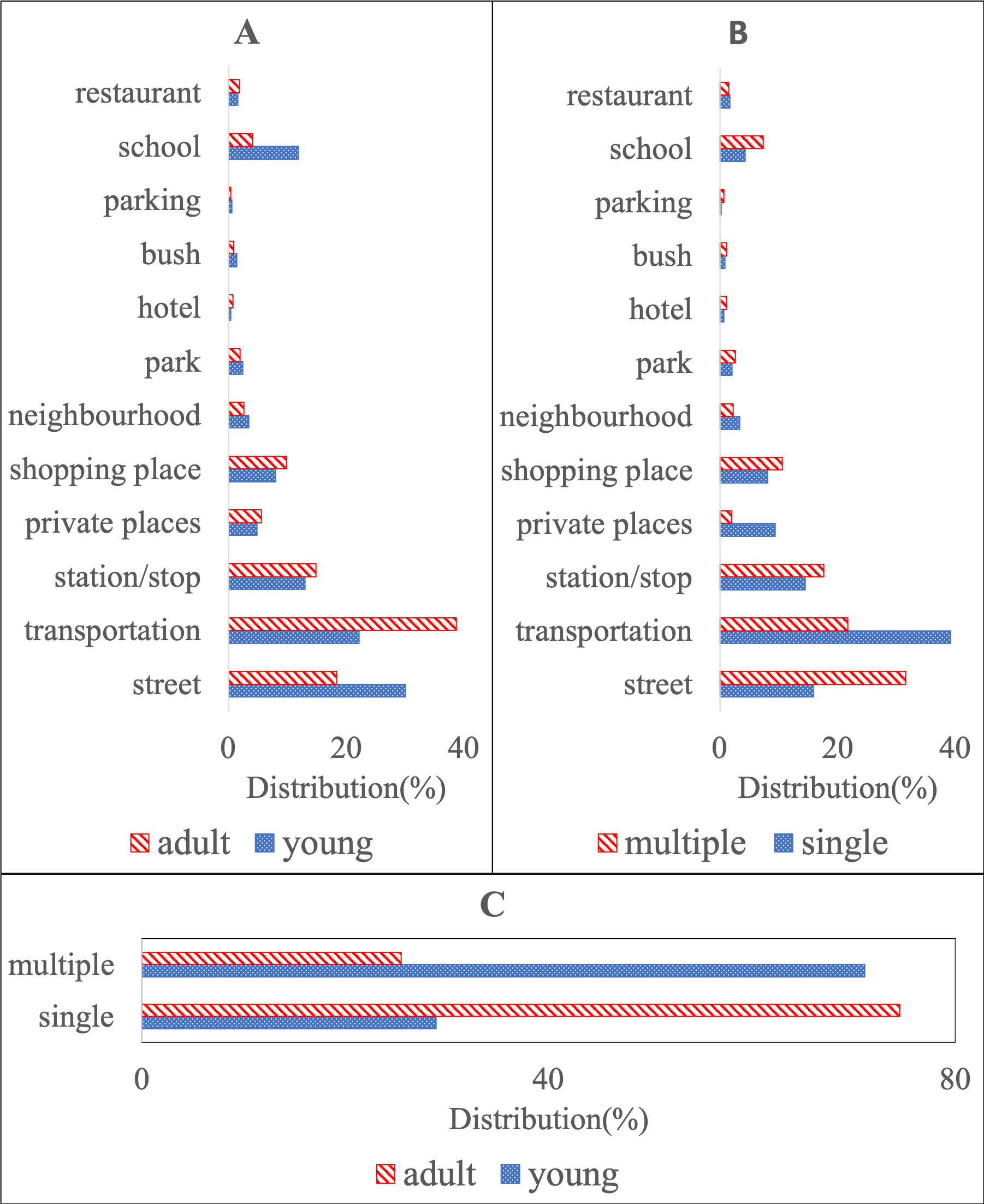}
    \caption{Distributions of incidents over two dimensions. A) Distributions of incidents A) with young/adult harassers at each location, B) with single/multiple harasser(s) at each location, C) across young/adult harassers and single/multiple harasser(s)} 
      \label{fig:correl}
          \vspace{-10pt}
\end{figure}

Furthermore, we uncovered that there exist strong correlations between the age of perpetrators and the location of harassment, between the single/multiple harasser(s) and location, and between age and single/multiple harasser(s) (Figure \ref{fig:correl}). The significance of the correlation is tested by chi-square independence with \textit{p} value less than 0.05. Identifying these patterns will enable interventions to be differentiated for and targeted at specific populations. For instance, the young harassers often engage in harassment activities as groups. This points to the influence of peer pressure and masculine behavioral norms for men and boys on these activities. We also found that the majority of young perpetrators engaged in harassment behaviors on the streets. These findings suggest that interventions with young men and boys, who are readily influenced by peers, might be most effective when education is done peer-to-peer. It also points to the locations where such efforts could be made, including both in schools and on the streets. In contrast, we found that adult perpetrators of sexual harassment are more likely to act alone. Most of the adult harassers engaged in harassment on public transportation. These differences in adult harassment activities and locations, mean that interventions should be responsive to these factors. For example, increasing the security measures on transit at key times and locations.

\begin{figure}[tb]
    \includegraphics[width=7.7 cm, height=6.8cm]{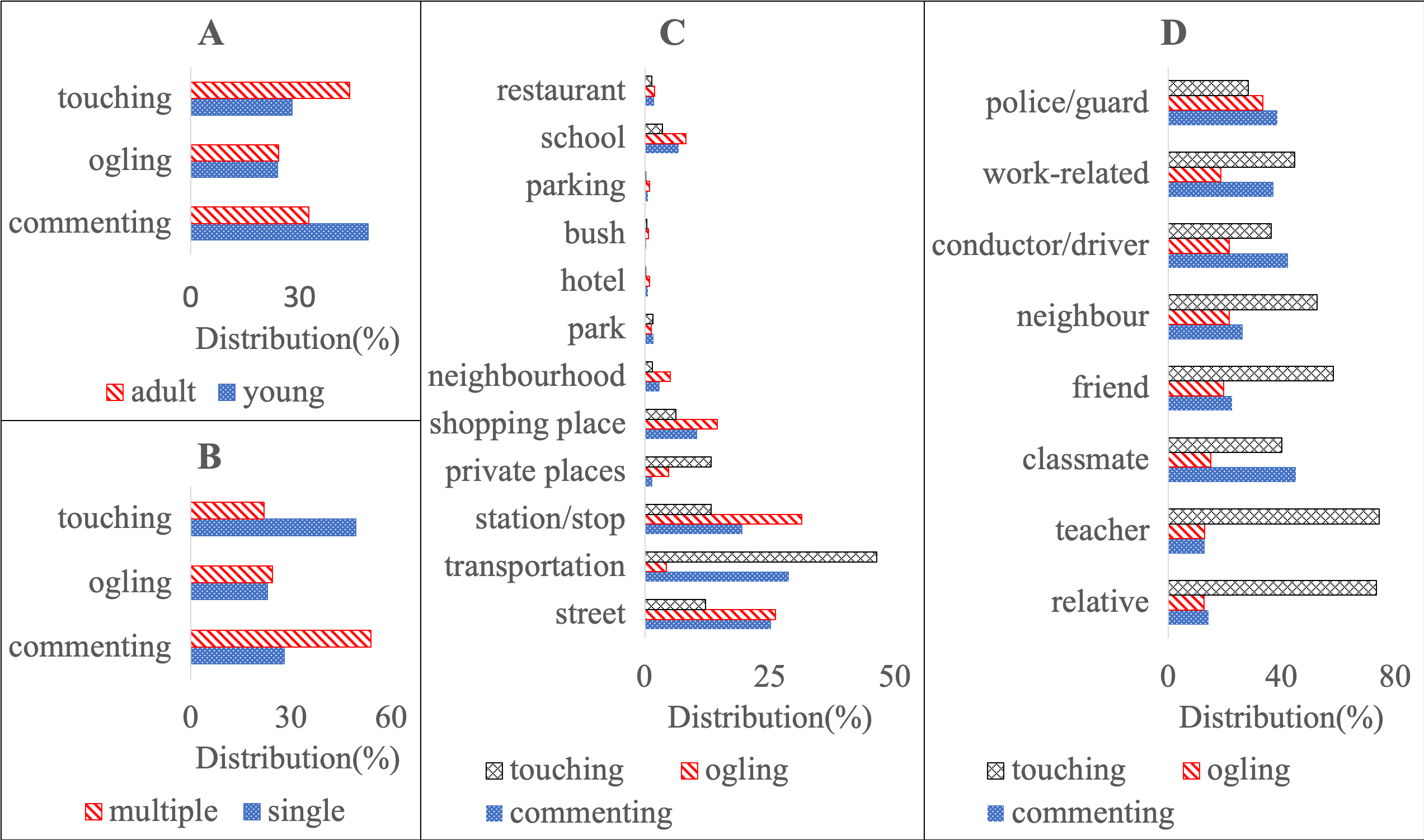}
  \caption{Distributions of incidents with harassment forms and different dimensions. Distributions of harassment forms A) within each age group, B) within single/multiple harasser(s), C) over locations, D) within each harasser type.}
    \label{fig:harasscorrel}
    \vspace{-10pt}
\end{figure}

In addition, we also found that the correlations between the forms of harassment with the age, single/multiple harasser, type of harasser, and location (Figure \ref{fig:harasscorrel}). For example, young harassers are more likely to engage in behaviors of verbal harassment, rather than physical harassment as compared to adults. It was a single perpetrator that engaged in touching or groping more often, rather than groups of perpetrators. In contrast, commenting happened more frequently when harassers were in groups. Last but not least, public transportation is where people got indecently touched most frequently both by fellow passengers and by conductors and drivers. The nature and location of the harassment are particularly significant in developing strategies for those who are harassed or who witness the harassment to respond and manage the everyday threat of harassment. For example, some strategies will work best on public transport, a particular closed, shared space setting, while other strategies might be more effective on the open space of the street.

These results can provide valuable information for all members of the public. Sharing stories of harassment has been found by researchers to shift people’s cognitive and emotional orientation towards their traumatic experiences \cite{Dimond2013}. Greater awareness of patterns and scale of harassment experiences promises to ensure those who have been subjected to this violence that they are not alone, empowering others to report incidents, and ensuring them that efforts are being made to prevent others from experiencing the same harassment. These results also provide various authorities tools to identify potential harassment patterns and to make more effective interventions to prevent further harassment incidents. For instance, the authorities can increase targeted educational efforts at youth and adults, and be guided in utilizing limited resources the most effectively to offer more safety measures, including policing and community-based responses. For example, focusing efforts on highly populated public transportation during the nighttime, when harassment is found to be most likely to occur.

\section{Conclusions}
We provided a large number of annotated personal stories of sexual harassment. Analyzing and identifying the social patterns of harassment behavior is essential to changing these patterns and social tolerance for them. We demonstrated the joint learning NLP models with strong performances to automatically extract key elements and categorize the stories. Potentiality, the approaches and models proposed in this study can be applied to sexual harassment stories from other sources, which can process and summarize the harassment stories and help those who have experienced harassment and authorities to work faster, such as by automatically filing reports \cite{karlekar2018safecity}. Furthermore, we discovered meaningful patterns in the situations where harassment commonly occurred. The volume of social media data is huge, and the more we can extract from these data, the more powerful we can be as part of the efforts to build a safer and more inclusive communities.  Our work can increase the understanding of sexual harassment in society, ease the processing of such incidents by advocates and officials, and most importantly, raise awareness of this urgent problem.
\section*{Acknowledgments}
We thank the Safecity for granting the permission of using the data.
\bibliography{EMNLP-2019-Long}
\bibliographystyle{acl_natbib}
\end{document}